\documentclass[conference]{IEEEtran}
\IEEEoverridecommandlockouts
\usepackage{cite}
\usepackage{amsmath,amssymb,amsfonts}
\usepackage{algorithmic}
\usepackage{graphicx}
\usepackage{textcomp}
\usepackage{xcolor}
\usepackage{url}            % simple URL typesetting
\usepackage{booktabs}       % professional-quality tables
\usepackage{amsfonts}       % blackboard math symbols
\usepackage{nicefrac}       % compact symbols for 1/2, etc.
\usepackage{microtype}      % microtypography
\usepackage{xcolor}         % colors
\usepackage{amsmath}    
\usepackage{subcaption}
\usepackage{caption}
\usepackage{cuted}
\usepackage{algorithmic}
\usepackage{algorithm} 
\usepackage{amsmath,amssymb,amsfonts}
\usepackage{placeins}
\def\BibTeX{{\rm B\kern-.05em{\sc i\kern-.025em b}\kern-.08em
    T\kern-.1667em\lower.7ex\hbox{E}\kern-.125emX}}
\begin{document}

\title{\textit{NOVA3D}: Normal Aligned Video Diffusion Model for Single Image to 3D Generation}

\author{
    \IEEEauthorblockN{1\textsuperscript{st} Yuxiao Yang$^{*}$, 2\textsuperscript{nd} Peihao Li$^{*}$ \thanks{$^{*}$ Equal Contributuion.}, 3\textsuperscript{rd} Yuhong Zhang, 4\textsuperscript{th} Junzhe Lu}
    
    \IEEEauthorblockA{\textit{Tsinghua University}, \textit{Tsinghua University}, \textit{Tsinghua University}, \textit{Tsinghua University}}
    
    \\
    
    \IEEEauthorblockN{5\textsuperscript{th} Xianglong He, 6\textsuperscript{th} Minghan Qin, 7\textsuperscript{th} Weitao Wang, 8\textsuperscript{th} Haoqian Wang$^{\dag}$ \thanks{$\dag$ Corresponding Author. wanghaoqian@tsinghua.edu.cn}}
    
    \IEEEauthorblockA{\textit{Tsinghua University}, \textit{Tsinghua University}, \textit{Tsinghua University}, \textit{Tsinghua University}} 
}

\maketitle

\begin{abstract}
    3D AI-generated content (AIGC) has made it increasingly accessible for anyone to become a 3D content creator. While recent methods leverage Score Distillation Sampling to distill 3D objects from pretrained image diffusion models, they often suffer from inadequate 3D priors, leading to insufficient multi-view consistency.
    In this work, we introduce \textit{NOVA3D}, an innovative single-image-to-3D generation framework. Our key insight lies in leveraging strong 3D priors from a pretrained video diffusion model and integrating geometric information during multi-view video fine-tuning. To facilitate information exchange between color and geometric domains, we propose the Geometry-Temporal Alignment (GTA) attention mechanism, thereby improving generalization and multi-view consistency.
    Moreover, we introduce the de-conflict geometry fusion algorithm, which improves texture fidelity by addressing multi-view inaccuracies and resolving discrepancies in pose alignment.
    Extensive experiments validate the superiority of \textit{NOVA3D} over existing baselines.


\end{abstract}

\vspace{-0.4cm}
\section{Introdution}    
    Creating 3D objects from a single-view image prompt is crucial for a wide range of applications in video games, virtual reality, and augmented reality. However, this task is highly ill-posed and presents significant challenges. Due to the difficulty in collecting high-quality 3D object data, 3D generative models\cite{jun2023shap, nichol2022point} lag behind their 2D counterparts in terms of realism and generalization. Therefore, leveraging prior information from related tasks, such as text-to-image generation, emerges as a promising approach to enhance both the realism and multi-view consistency of generated 3D objects.
    
    A growing body of works\cite{poole2022dreamfusion,  lin2023magic3d} resort to distilling a 3D representation from a pretrained text-to-image model via Score Distillation Sampling(SDS)\cite{poole2022dreamfusion}. To enhance both multi-view consistency and efficiency, an alternative approach\cite{liu2023syncdreamer, liu2023zero, long2023wonder3d, shi2023mvdream} utilizes image diffusion models fine-tuned on 3D datasets\cite{deitke2023objaverse} for multi-view image generation, followed by a reconstruction process to derive a 3D object. 
    Although these methods alleviate the extensive high-quality 3D data requirements, they suffer from blurry back-view texture, insufficient generalizability, and limited 3D consistency.
    Humans, by contrast, derive 3D priors primarily from dynamic observations (e.g. videos), which allow for the inference of 3D structures from a single image. 
    Inspired by this capability, there is substantial potential to explore and exploit 3D priors embedded in large-scale pretrained video models to enhance single-image-to-3D generation.
    
    % 这一段讲video diffusion和video diffusion和几何的关系
    Recent advancements in video diffusion models\cite{blattmann2023stable, blattmann2023align} have garnered considerable attention for their remarkable capability to generate intricate scenes and complex dynamics with exceptional cross-frame consistency. While some studies have employed fine-tuned video diffusion models to generate multi-view images\cite{pang2024envision3d, chen2024v3d, voleti2024sv3d}, the potential of video diffusion models for capturing and understanding 3D geometry remains under-explored. Consequently, these methods often struggle to model detailed geometric structures and produce high-fidelity texture details.

    \begin{figure*}[!t]
\centering
\includegraphics[width=0.9\linewidth]{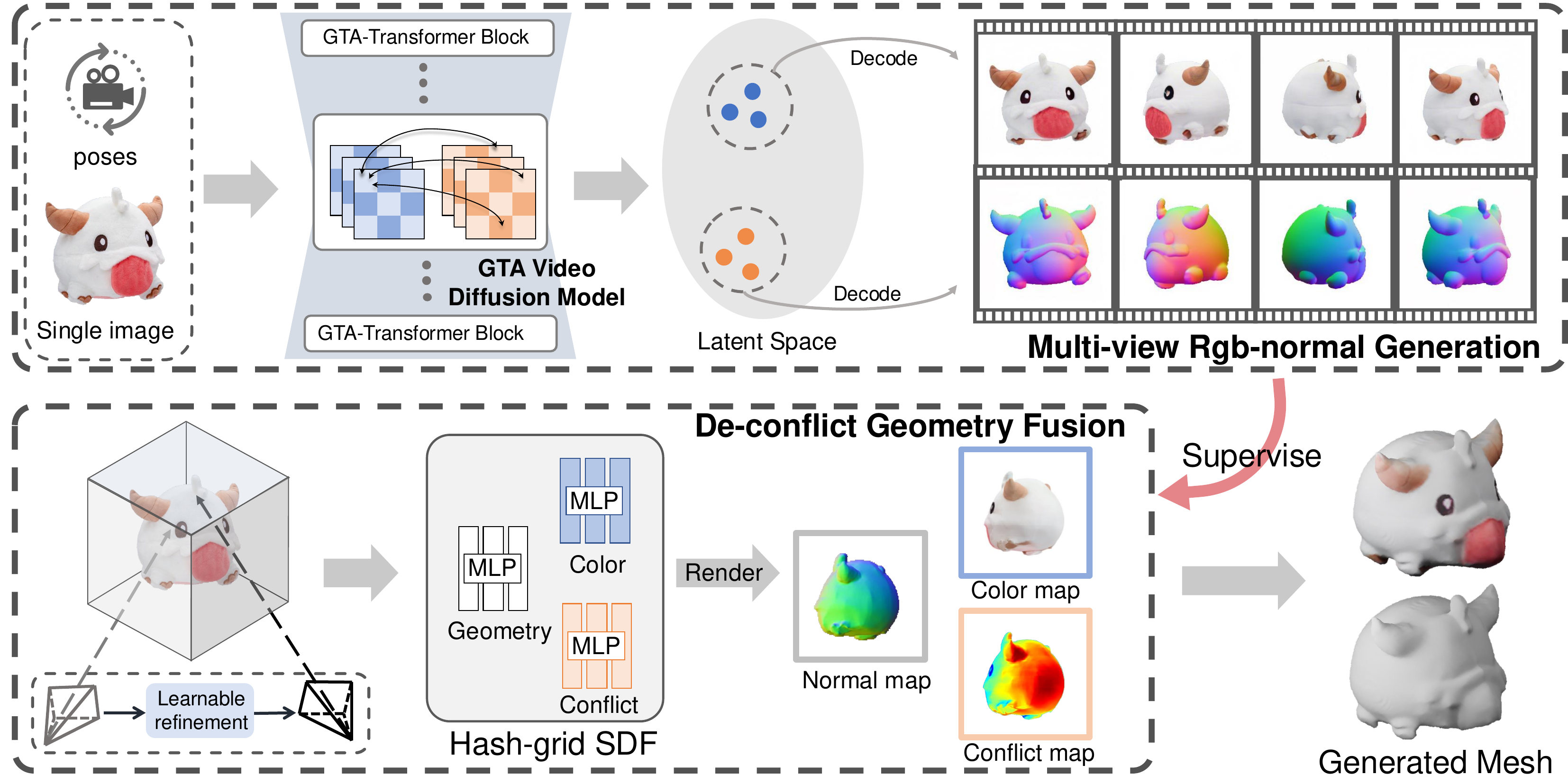}

\caption{
\textbf{Overview of the \textit{NOVA3D} pipeline.} Our approach starts by leveraging a GTA-infused video diffusion model to generate multi-view images and their corresponding normal maps from a single image. These results are subsequently processed through a de-conflict geometry fusion algorithm to reconstruct a high-fidelity textured mesh that accurately captures the details.
}
\vspace{-0.6cm}
\label{fig:pipeline}
\end{figure*}
    To enhance multi-view consistency and fully leverage geometric priors from pre-trained video diffusion models, in this paper, we introduce \textit{NOVA3D}, a novel framework that utilizes 3D priors embedded in pre-trained video diffusion models to generate high-quality textured meshes from single-view images. Our key insight lies in incorporating geometric information as auxiliary supervision, which augments the activation of 3D priors within the pretrained video diffusion model. This refinement empowers the video diffusion model to predict multiview images and corresponding normal maps, consequently facilitating the reconstruction of high-fidelity textured meshes. Moreover, we introduce the innovative Geometry-Temporal Alignment (GTA) attention mechanism into the Latent Video Diffusion Model(LVDM) architecture, which aligns the generation of RGB images and normal maps, thereby migrating generalizability from RGB video domain to the geometric domain without modifying the pre-trained model. 
    To address discrepancies between generated and predefined poses, as well as subtle cross-view inconsistencies, we present the de-conflict geometry fusion algorithm. This algorithm incorporates implicit conflict modeling and pose refinement techniques, ensuring robust and consistent textured mesh generation.
    Our evaluation on both the Google Scanned Object dataset\cite{downs2022google} and out-of-distribution inputs demonstrates the efficacy of \textit{NOVA3D}, with quantitative results indicating superior fidelity and generalizability compared to baseline methods.

    To sum up, our contribution can be summarized as follows:
    
    \begin{itemize}
    \item We introduce \textit{NOVA3D}, a novel approach unleashing geometric 3D prior from a video diffusion model to generate high-quality textured meshes from input images.
    
    \item We propose the Geometry-Temporal Alignment attention mechanism to facilitate the exchange of patterns between texture and geometric latents, effectively transferring generalization performance to the geometric domain.

    \item We present a de-conflict geometry fusion algorithm, incorporating implicit conflict modeling and pose refinement techniques, improving the robustness and texture fidelity.
    \end{itemize}

\section{Related Works}

\subsection{Image Diffusion Models for 3D Generation}

    In recent years, image diffusion models \cite{ho2020denoising, song2020denoising, rombach2022high} have seen rapid development. 
    However, the relative scarcity of 3D data limits the performance of native 3D generation models.
    Previous works have attempted to leverage pretrained image diffusion models for 3D object generation. 
    For instance, DreamFusion \cite{poole2022dreamfusion} proposed the SDS method for Text-to-3D tasks, optimizing a neural radiance field \cite{mildenhall2021nerf} guided by textual prompts. Although subsequent studies \cite{wang2024prolificdreamer, lin2023magic3d} have focused on improving SDS through multi-stage optimization, enhanced distillation, and accelerated distillation methods,  the time-consuming per-object optimization and the multi-face problem still hinder this approach impractical for real-world applications.
    To address this, recent methods\cite{long2023wonder3d, liu2023syncdreamer} fine-tune image diffusion models on 3D datasets \cite{deitke2023objaverse}, enabling the generation of multi-view images consistent with the input. 
    Nonetheless, due to the relative lack of 3D priors, these models often need to train cross-view attention layers from scratch on 3D datasets to ensure multi-view consistency, which hampers their ability to generate high-quality, dense-view images.

\subsection{Video Diffusion Model for 3D Generation}

    More recently, research on video diffusion models \cite{blattmann2023align, blattmann2023stable} has progressed significantly. Pretraining generative models on massive real video datasets
    % , e.g. WebVid-10M\cite{bain2021frozen} 
    provide extensive 3D prior, including object interactions, rotations, and camera movements. Some recent works have attempted 3D object generation using prior from video diffusion models\cite{voleti2024sv3d, chen2024v3d}, treating multi-view images of objects as sequential frames and fine-tuning video diffusion models using rendered multi-view images from 3D datasets. 
    However, existing methods have not fully exploited the 3D information within video diffusion models. 
    Therefore, we propose to incorporate geometric information as supervision alongside texture information to fine-tune the video diffusion model, effectively activating 3D prior information within the video diffusion models.

\section{Method}

\subsection{Problem Formulation.}

    Given an input image of an object $y$ and a series of pre-defined camera poses $\pi_{i:m}$, there exists a probabilistic distribution of m-views of the color images and normal maps:
    \begin{equation}
    p_{ni}(i_{1:m}, n_{1:m} | y, \pi_{i:m}).
    \end{equation} 
   Our goal is first to sample $m$ views of multiview images $i_{i:m}$ and corresponding normal maps $n_{1:m}$ from the distribution of $p_{ni}$, and then perform the de-conflict geometry fusion algorithm to generate a textured mesh. 
   We assume that the object is located at the center of the normalized 3D cube and adopts a series of camera poses evenly distributed at an elevation angle of $0$, thus removing the need to input the elevation angle. 
   Specifically, we generate multi-view images and normal maps that match the following distribution:
   \begin{equation}
        i_{1:m}, n_{1:m} = f(y, \pi_{1:m}) \sim p(i_{1:m}, n_{1:m} | y, \pi_{i:m})
    \end{equation}
    where $f$ is our fine-tuned video diffusion model. 

    \subsection{Unleashing the 3D priors within video diffusion model.} \label{sec: Video Diffusion Model for Cross-Domain Multiview Generation}
    
    \noindent
    \textbf{Overall Architecture.} By introducing a temporal dimension, a Conv3D residual layer, and a temporal attention layer after each spatial layer, the latent video diffusion model\cite{blattmann2023align} generates a temporally consistent sequence of images. \textit{NOVA3D} adopts this architecture and initializes the weights from SVD\cite{blattmann2023stable}, ensuring temporal consistency and providing a strong prior for multi-view generation. 
    % At noise timestep $t$, the VAE encoder of SVD embeds the conditioning image, which is then concatenated with the noisy latent $z_t$. 
    The CLIP embedding of the conditioning image is subsequently used as key and value in the cross-attention layers of the transformer blocks within the video UNet. 
    We made several adjustments to make the pre-trained model suitable for our task: (a) removing 'motion bucket id' and 'fps id' inputs as they are irrelevant for multi-view generation; (b) integrating camera conditioning $\pi_i$ and one-hot encoded task conditioning $t_i$. These conditions, embedded as labels in the video U-net for each sequence, represent the query pose and determine to generate appearance or geometry. 

    \noindent
    
\begin{figure}[t!]
    \centering
    % \vspace{-0.3cm}
    \includegraphics[width=0.45\textwidth]{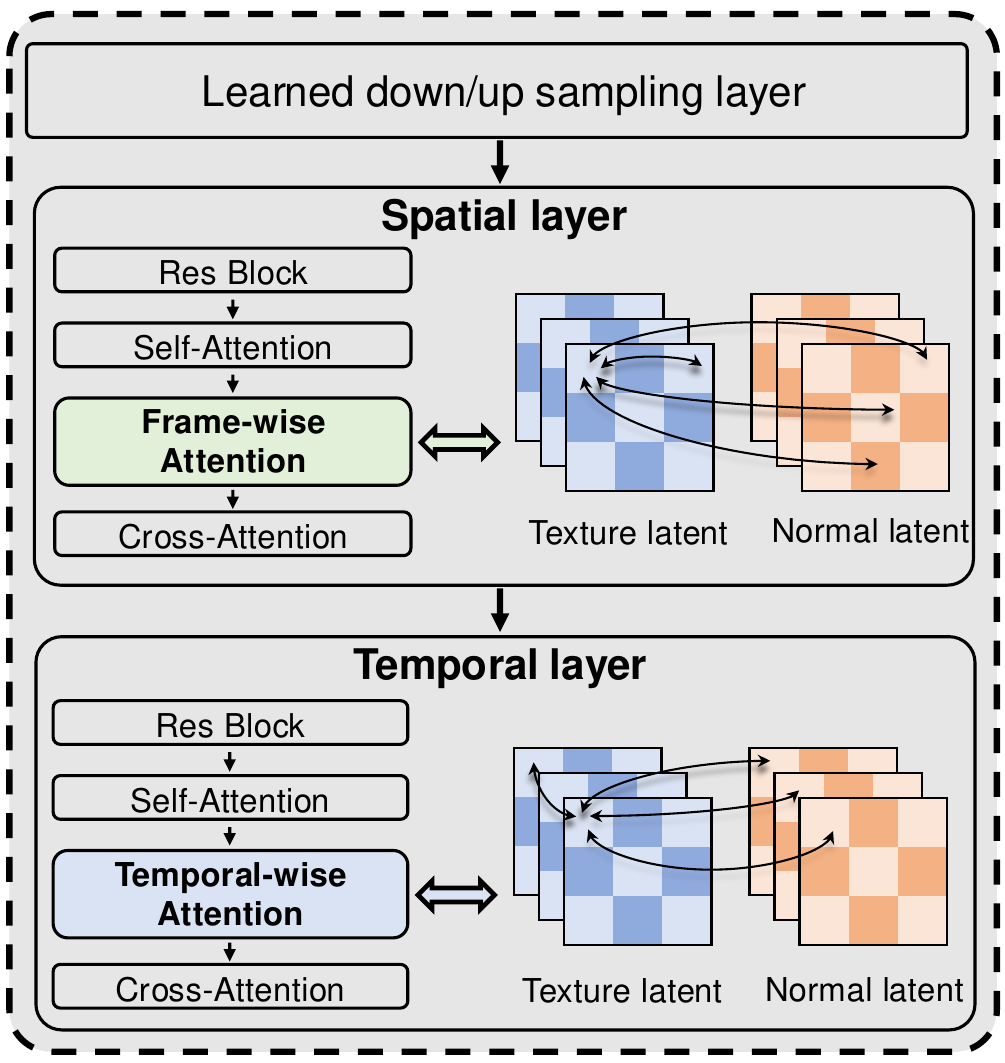}
    \caption{\textbf{Illustration of GTA attention mechanism.} The proposed GTA attention mechanism ensures efficient interaction between texture and geometry features at each spatial and temporal layer within LVDM. }
    \label{fig:GTA}
    \vspace{-0.6cm}
\end{figure}
    \noindent
    \textbf{Incorporation of Geometry.} For the task of multi-view image generation tasks, previous works\cite{voleti2024sv3d, chen2024v3d} have highlighted the generalization of video diffusion models fine-tuned on multi-view images rendered from 3D object datasets. 
    To exploit 3D priors within the pretrained SVD, we integrate geometric information during the finetuning of our model.
    To achieve this, straightforward approaches typically involve either doubling the channels within the U-net or first generating a sequence of images and then conditioning them to produce the corresponding normal maps. However, both methods necessitate weight reinitialization, causing catastrophic forgetting of the model and reducing generalization performance.

    Unlike the approaches mentioned above, our method offers control over the model's output task, allowing a seamless transition between color and geometry domains through the task condition. This enhancement not only obviates the need for U-Net parameter reinitialization but also leverages geometric information as an additional constraint, thereby enhancing the 3D prior knowledge ingrained during pre-training. The rationale behind this design is that multi-view color images often lack sufficient information to accurately reflect the true 3D structure of objects, especially for textureless surfaces. Therefore, the supervision of normal maps serves as an additional constraint, facilitating a smoother adaptation of the video diffusion model from a video-generation task to a multi-view generation task. 
    % See Sec\ref{Diccusion} for a more detailed 
    % transition that our experiments will aptly illustrate.
    % We train NOVA3D on the LVIS subset of Objaverse dataset\cite{deitke2023objaverse}, which contains around $30000$ objects.

\subsection{Geometry-Temporal Alignment Attention Mechanisim}

    \noindent
    \textbf{Multi-task Denoise Procedure.} The enhancements we introduced in Section \ref{sec: Video Diffusion Model for Cross-Domain Multiview Generation} empower our model to generate multi-view images and normal maps without a significant modification to the network. While the RGB images and normal maps each ensure consistency across views, directly meshing leveraging them may result in a misalignment between texture and geometry. Furthermore, there is an interconnection between the color and geometry of an object. Therefore, considering both at the same time will help the model learn the true distribution of a 3D object. We formulate our denoise process as follows:
    \begin{equation}
    \begin{aligned}
    p(i_{1:m}, n_{1:m}| \pi_{1:m}, y) &= p(i_{1:m}^{T}, n_{1:m}^{T} | \pi_{1:m}, y) \\
    & \hspace{-6em} \cdot \prod_{t \in 1:T} p_{\theta} (i_{1:m}^{t-1}, n_{1:m}^{t-1} | i_{1:m}^{t}, n_{1:m}^{t}, \pi_{1:m}, y).
    \end{aligned}
    \end{equation}
    This indicates that at each denoise step $t$, our model $f$ is performed as a noise predictor that predicts noise on the noised multi-view color images $i^t_{1:m}$ and corresponding normal maps $n^t_{1:m}$ to derive the de-noised result $i^{t-1}_{1:m}$ and $n^{t-1}_{1:m}$ jointly. 

    \noindent
    \textbf{GTA Attention Module.} To enable the model to generate aligned color and normal maps and facilitate the pattern exchange between texture and geometry domains, we propose the Geometry-Temporal Alignment (GTA) attention mechanism. 
    Figure \ref{fig:GTA} illustrates the operational dynamics of the GTA attention mechanism. Specifically, at the spatial level, the GTA attention mechanism enables efficient interaction between RGB images and normal maps within the same viewpoint. Simultaneously, at the temporal level, it ensures alignment across different viewpoints at corresponding positions within the latent feature map. This streamlined approach harmonizes with the intricate information processing patterns embedded within the latent video diffusion model architecture. See supplementary for more implementation details.

\subsection{De-conflict Geometry Fusion Algorithm}
    \begin{figure*}[t]
\centering
\includegraphics[width=0.85\linewidth]{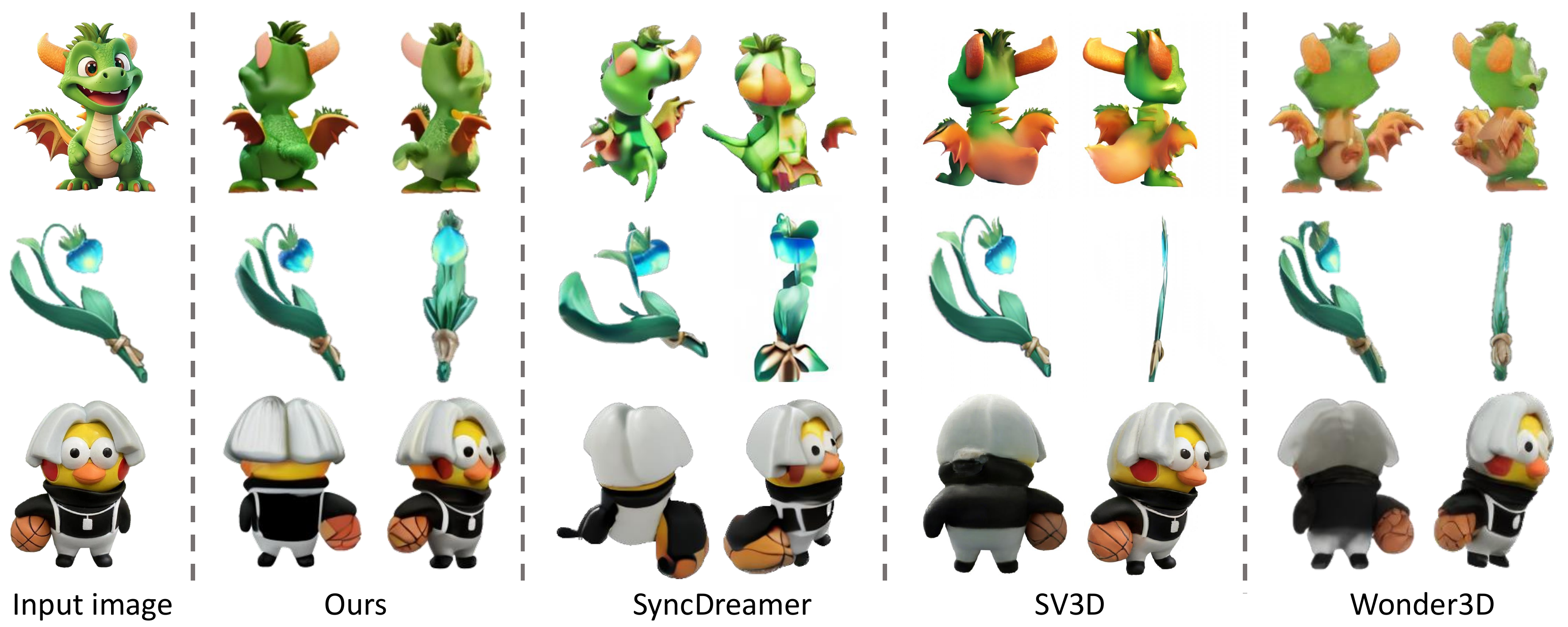}
\vspace{-0.1cm}
\caption{
     \textbf{Qualitative results of novel view synthesis on out-of-distribution images.}
}
\vspace{-0.4cm}
\label{fig:nvs}
% \vspace{-.1in}
\end{figure*}
    \begin{table}[!t]
\centering
\resizebox{0.95\linewidth}{!}{

\begin{tabular}{lccccc}
\toprule
\multicolumn{1}{c}{} & \multicolumn{2}{c}{Mesh Reconstruction} & \multicolumn{3}{c}{Texture Quality} \\
\cmidrule(lr){2-3}
\cmidrule(lr){4-6}
 \multicolumn{1}{l}{Methods} & $\downarrow$Chamfer Dist. & $\uparrow$Volume IoU  & $\uparrow$PSNR & $\uparrow$SSIM & $\downarrow$LPIPS \\
\midrule
One-2-3-45~\cite{liu2023zero}    & 0.0629          & 0.409          & -          & -          & -   \\
Shape-E ~\cite{jun2023shap}  & 0.0436          & 0.358          & -          & -          & -          \\
Zero123~\cite{liu2023zero}    & 0.0339          & 0.504          & 16.60          & 0.798          & 0.207          \\
SyncDreamer~\cite{liu2023syncdreamer} & 0.0261          & 0.542          & 16.02          & 0.770          & 0.249          \\
V3D~\cite{chen2024v3d} & 0.0250          & 0.552          & 15.52         & 0.780          & 0.227          \\
Wonder3D*~\cite{long2023wonder3d}   & 0.0242          & 0.578  & 15.89          & 0.784          & 0.214         \\
Envision3d~\cite{pang2024envision3d}  & 0.0238  &  0.593  & 20.00  & 0.845  & 0.165  \\
CRM ~\cite{wang2024crm} & 0.0225  & 0.598 & 21.74  & 0.862  & 0.155  \\
Ours        & \textbf{0.0212}  & \textbf{0.602}   & \textbf{22.11} & \textbf{0.872} & \textbf{0.126}\\
\bottomrule
\end{tabular}

}
\vspace{-1mm}
\caption{\textbf{Quantitative results on mesh reconstruction and re-render views.} 
To compare the quality of texture, we additionally report PSNR, SSIM, and LPIPS of the re-rendered images.}
\vspace{-0.6cm}
\label{tab:mesh_rerender}
\end{table}
     Incorporating our generated normal maps to aid in 3D geometry and texture extraction, we employ an implicit signed distance function during optimization, thereby simplifying the computation of normal map loss. 
     Regrettably, our reconstruction process faces two potential challenges: (a) minor deviations between the generated pose and query poses, and (b) subtle inconsistencies among the overlapping views due to the relatively dense nature of the 16-views generation. To mitigate these challenges, we introduce the \textbf{de-conflict geometry fusion} algorithm, which we will discuss as follows.

     \noindent
     \textbf{Pose Refinement.} In order to address the misalignment between the generated pose and the pre-defined query pose, we introduce a pose refinement technique. 
     Initially set according to query poses, camera poses $\pi_{1:m}$, represented by rotation matrix and translation vectors for each view, undergo refinement during optimization. 
     Specifically, each ray starting from the $v_{th}$ view shifts via a learnable refinement matrix $M_v$, which remains consistent across all rays within the same view. This approach refines poses to the correct angles, enhancing the quality of the generated mesh.
    
     \noindent
     \textbf{Conflict Modeling.} Subtle conflicts between adjacent views contribute to optimization instability, resulting in blurred geometry and texture. To tackle this, we employ an implicit continuous function $f_{\psi}$ to model conflicts $h$ between overlapping images:
     \begin{equation}
         h = f_{\psi} (f_c, f_g, d(v), l, x)
     \end{equation}
     where $f_c$, $f_g$, $d(v)$, $l$, and $x$ denote the output of the color MLP, geometry MLP, ray direction, view index embedding, and coordinate position, respectively. 
     Conflicts at pixel $p$ in the camera space, denoted as $H_p$, are computed by projecting the ray’s direction onto the 2D-pixel plane via volume rendering.  $H_p$  quantifies the conflict of pixel $p$ with adjacent images.
     Our de-conflict color loss is defined as:
     \begin{equation}
         \mathcal{L}_{color} = (1 - H_p) \Vert C_p - \hat{C_p} \Vert_2 + \lambda_{0} H_p^2
     \label{eq:rgb_loss}
     \end{equation}
     where $C_p$ and $\hat{C}_p$ are the rendered pixel colors and the generated image colors. In this equation, pixels with higher conflict values are given smaller weights, thus reducing the negative impact of inconsistency between overlapping views during the reconstruction. The second equation serves as a regularization term, preventing $H_p$ from becoming excessively large, which would undermine the color supervision signal.

     \noindent
     \textbf{Loss Function.} 
     \begin{table}[!t]
    \centering
    \begin{tabular}{lccc}
        \toprule
        Methods & $\uparrow$PSNR & $\uparrow$SSIM & $\downarrow$LPIPS  \\
        \midrule
        Zero123~\cite{liu2023zero} & 18.93 & 0.779 & 0.166 \\
        SyncDreamer~\cite{liu2023syncdreamer} & 20.05 & 0.798 & 0.146 \\
        V3D~\cite{chen2024v3d} & 20.22 & 0.812 & 0.132 \\
        Envision3D~\cite{pang2024envision3d} & 20.55 & 0.852 & 0.130 \\
        SV3D~\cite{voleti2024sv3d} & 20.88 & 0.897 & 0.112 \\
        % Era3D ~\shortcite{li2024era3d} & 23.13 & 0.810 & 0.126 \\
        Wonder3D~\cite{long2023wonder3d} & 23.25 & 0.822 & 0.104 \\

        \midrule
        w/o GTA & 22.10 & 0.802 & 0.144 \\
        w/ cross-domain attn. & 23.26 & 0.824 & 0.108 \\
        Ours & \textbf{23.87} & \textbf{0.915} & \textbf{0.091} \\
        \bottomrule
    \end{tabular}
    \caption{\textbf{Quantitative results in novel view synthesis.}}
    \label{tab:nvs_quantative}
    \vspace{-0.4cm}
\end{table}

    We sample a batch of rays for each iteration of the optimization process. Given a point $k$ on the ray, we query the geometry, color, and conflict MLPs to render it along the ray direction to derive normal map value $h_k \in \mathbb{R}$, the color value $c_k$, and the mask $m_k$. 
    \begin{figure*}[!t]
\centering
\vspace{-6mm}
\includegraphics[width=0.9\linewidth]{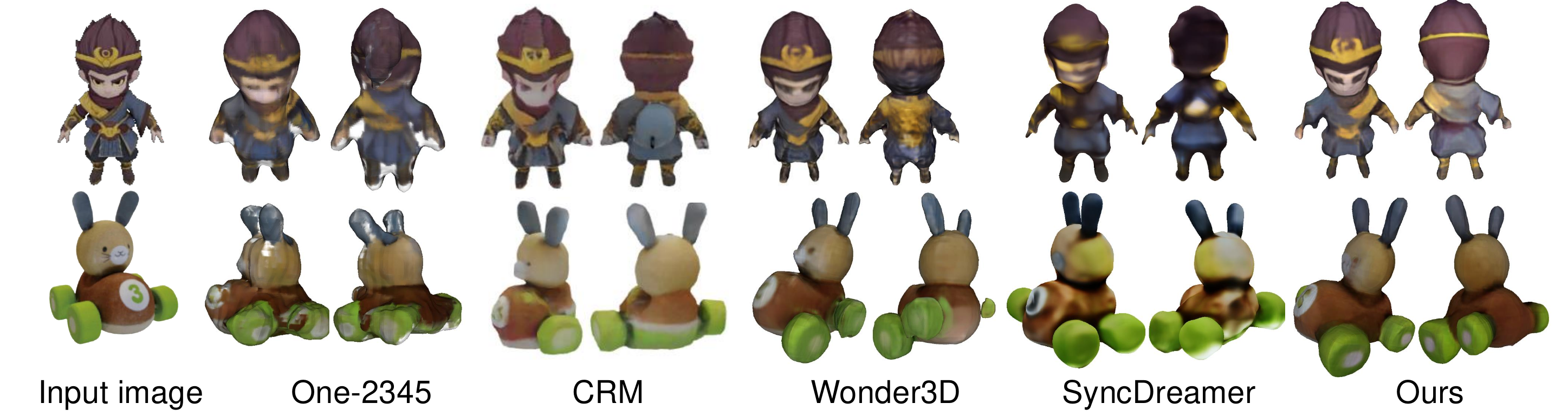}
\caption{
     \textbf{Qualitative comparison with baselines in terms of the generated textured meshes.}
}
\label{fig:mesh_vis}
% \vspace{-.1in}
\vspace{-2mm}
\end{figure*}
    
    The final optimization objective integrates multiple loss terms:
     \begin{equation}
         \mathcal{L} = \mathcal{L}_{color} + \mathcal{L}_{normal} + \mathcal{L}_{mask} + \mathcal{R}_{eik} + \mathcal{R}_{sparse} + \mathcal{R}_{smooth}
     \end{equation}
     where $\mathcal{L}_{color}$ is our de-conflict loss mentioned above, $\mathcal{L}_{normal}$ is the Geometry-aware Normal Loss proposed in Wonder3D\cite{long2023wonder3d}, which maximizes the similarity of generated normal and the extracted normal value from SDF representing, $\mathcal{L}_{mask}$ is a L2 loss between rendered mask $m_k$ and generated mask $\hat{m}_k$, $\mathcal{R}_{eik}$\cite{gropp2020implicit}, $\mathcal{R}_{sparse}$\cite{long2022sparseneus} and $R_{smooth}$\cite{long2023wonder3d} are regularization terms aimed at enforcing the predicted SDF to have a unit $l_2$ norm gradient, avoiding floaters, and encouraging smoother predicted SDF gradients, respectively.

\section {Experiments}

\subsection {Implementation Details}
    \begin{figure}
\centering
\vspace{-.2cm}
\includegraphics[width=0.9\linewidth]{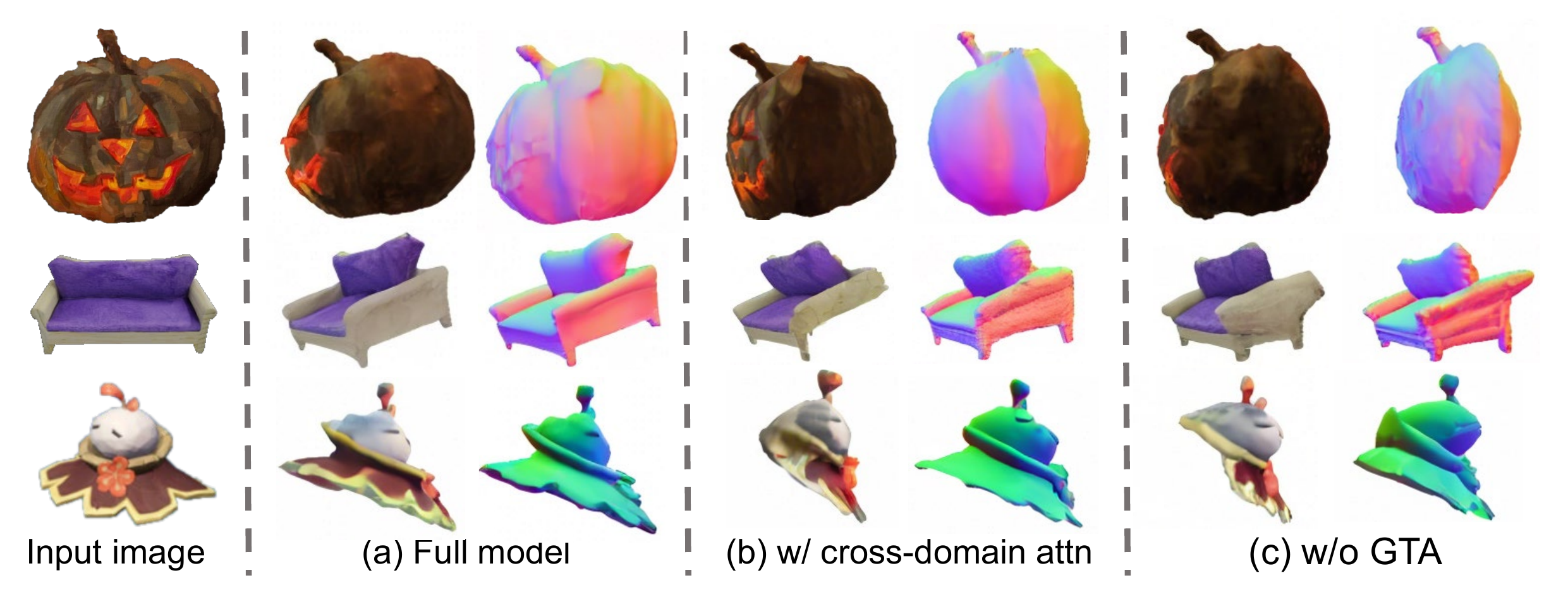}
\caption{
     \textbf{Ablation studies of GTA attention mechanism.}
}
\vspace{-0.4cm}
\label{fig:ablation}
% \vspace{-.1in}
\end{figure}
    
    We conduct the training on the LVIS subset of the Objaverse dataset\cite{deitke2023objaverse}, which comprises approximately 30,000 3D meshes. 
    RGB images and normal maps are rendered at 16 poses, each at a resolution of $256 \times 256$,  for training our model.

    Our model is fine-tuned from the publicly available SVD\cite{blattmann2023stable} on 8 Nvidia A100 GPUs over a period of 7 days with an effective batch size of 176. 
    During SDF optimization, we use the hierarchical hash grid\cite{muller2022instant} to encode 3D positions with multi-level detail, improving efficiency.

\subsection {Evalutation Settings}
    \noindent
    \textbf{Baselines.} We evaluate our method against several single image to 3D approaches, including Zero123\cite{liu2023zero}, SyncDreamer\cite{liu2023syncdreamer}, Wonder3D\cite{long2023wonder3d}, as well as recent video diffusion model-based methods such as Envision3D\cite{pang2024envision3d}, V3D\cite{chen2024v3d}, and SV3D\cite{voleti2024sv3d}. 
    In addition, we perform a comparative analysis with various feedforward 3D generative approaches, including Shap-E \cite{jun2023shap}, One-2-3-45 \cite{liu2024one}, and CRM \cite{wang2024crm}. This comprehensive evaluation demonstrates the effectiveness and robustness of our method across diverse benchmarks and scenarios.
    
    \noindent
    \textbf{Metrics.} Following prior work\cite{liu2023syncdreamer, long2023wonder3d}, we evaluate our method on the Google Scanned Object\cite{downs2022google} dataset, selecting 30 objects ranging from daily items to animals. 
    % Our evaluation focuses on both novel view synthesis (NVS) and mesh reconstruction. 
    For the NVS task, we use PSNR, SSIM\cite{wang2004image}, and LPIPS\cite{zhang2018unreasonable} metrics to assess the quality of our generated multi-view images. 
    To evaluate the quality of our generated textured meshes, we first adopt Chamfer Distance and Volume IoU metrics for geometry evaluation. Additionally, we re-render generated meshes at 32 fixed poses, as utilized by Envision3D\cite{pang2024envision3d}, to evaluate the quality of the mesh textures.

\subsection {Novel View Synthesis}

    We evaluate the quality of the generated multi-view images, presenting qualitative results in Fig. \ref{fig:nvs} and quantitative results in Table \ref{tab:nvs_quantative}. 
    The outputs of SyncDreamer\cite{liu2023syncdreamer} lack multi-view consistency and exhibit unrealistic artifacts.
    Wonder3D\cite{long2023wonder3d} employs a multi-view attention mechanism that achieves relatively consistent multi-view images but is limited to generating only six views, significantly fewer than the sixteen views generated by our approach. While SV3D\cite{voleti2024sv3d} achieves view consistency by fine-tuning SVD with RGB-only information, the overall shape realism and color detail of the generated objects are insufficient In contrast, our model, supported by auxiliary supervision from geometric information and leveraging essential 3D priors within pretrained SVD, excels in producing multi-view images that are both consistent across views and semantically coherent.

\vspace{-1mm}    
\subsection {Textured Mesh Generation}
\vspace{-1mm}    
    \begin{table}[!t] 
\centering
\small
\resizebox{1\linewidth}{!}{ 
\begin{tabular}{lccccc}
\toprule
\multicolumn{1}{c}{} & \multicolumn{2}{c}{Mesh Reconstruction} & \multicolumn{3}{c}{Texture Quality} \\
\cmidrule(lr){2-3}
\cmidrule(lr){4-6}
\multicolumn{1}{l}{Methods} & $\downarrow$Chamfer Dist. & $\uparrow$Volume IoU  & $\uparrow$PSNR & $\uparrow$SSIM & $\downarrow$LPIPS \\
\midrule
w/o GTA  & 0.0427     & 0.3197          & 18.72          & 0.835            & 0.156  \\
w/ cross-domain atten. & 0.0256          & 0.5432           & 19.44          & 0.846          & 0.144      \\
w/o conflict  & 0.0244       & 0.5932          & 21.68          & 0.855          & 0.136    \\
w/o pose-refine    & 0.0232          & 0.5916         & 21.89          & 0.861          & 0.132      \\
Ours        & \textbf{0.0212} & \textbf{0.6021}          & \textbf{22.11} & \textbf{0.872} & \textbf{0.126}\\
\bottomrule
\end{tabular}
}
\caption{\textbf{Ablation studies on mesh reconstruction.}}
\vspace{-6mm}
\label{tab:aba_mesh}
\end{table}
    We evaluate and compare both the geometry and texture quality of the generated meshes against state-of-the-art methods. As shown in Table \ref{tab:mesh_rerender}, our method demonstrates superior performance across all metrics, highlighting its capability to produce high-fidelity 3D content with rich texture details. Fig \ref{fig:mesh_vis} presents qualitative comparisons of the generated textured meshes, further illustrating that our method significantly surpasses baselines in terms of mesh geometry, texture, and high-level semantic consistency.

\subsection {Disscusion}
\label{Diccusion}

    \noindent
    \textbf{Geometry-Temporal Alignment (GTA) Attention.} To validate the effectiveness of the proposed GTA attention mechanism, we conducted experiments with different model configurations: (a) finetuning a video diffusion model incorporating the GTA attention module, (b) finetuning utilizing the cross-domain attention module as proposed by Wonder3D\cite{long2023wonder3d}, and (c) a variant model without either the GTA or cross-domain attention modules. Figure~\ref{fig:ablation} shows the visualizations, while Tables~\ref{tab:nvs_quantative} and \ref{tab:aba_mesh} present the quantitative results.

    Comparing (a) and (b) in Figure \ref{fig:ablation}, the lack of GTA attention in (b) hampers the exchange of information between frames, resulting in geometric normals that fail to comprehend the overall shape of the object and align with the generated texture information.
    Similarly, as depicted in (a) and (c) of Figure \ref{fig:ablation}, the lack of interaction between color and normal features impedes the transfer of generalizability from the texture domain to the geometric domain.
    In contrast, the integration of the GTA module within the video diffusion model architecture, as shown in (a), enables the generation of sharper and more accurate normal maps while ensuring superior consistency between color and normal features. This highlights the effectiveness of our approach in bridging multi-task and multi-view dependencies.
    
\begin{figure}[tp!]
\centering
\vspace{-0.4cm}
\includegraphics[width=0.88\linewidth]{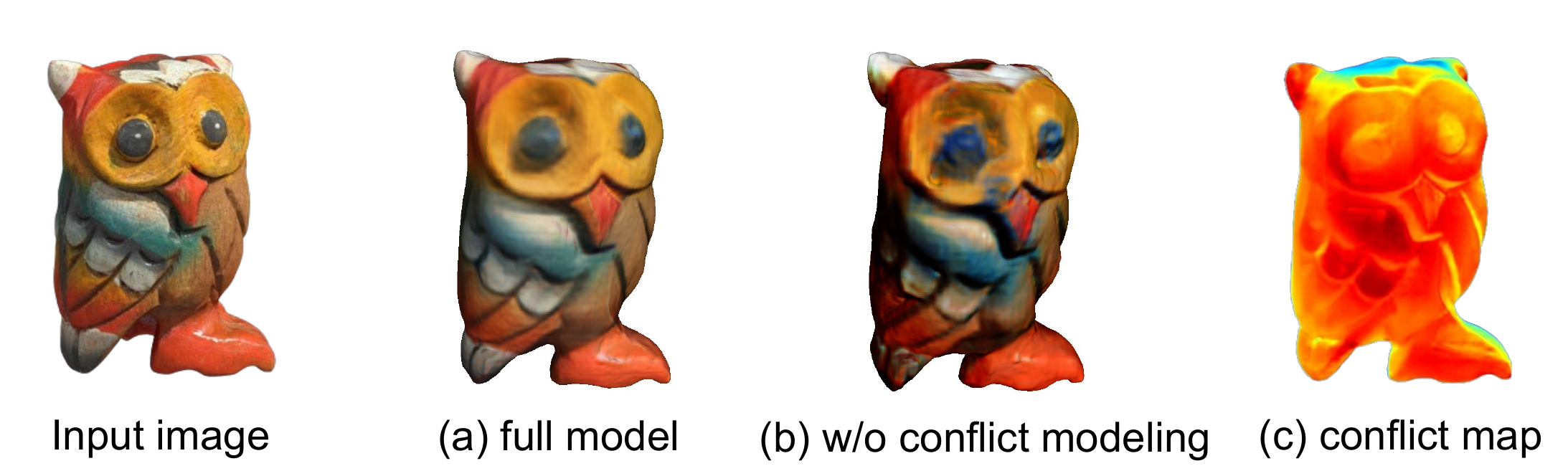}
\caption{\textbf{Ablation study on implicit conflict modeling.}}
\label{fig:ablations_recon}
\vspace{-0.6cm}
\end{figure}

    \noindent
    \textbf{De-conflict Geometry Fusion Algorithm.} We evaluated the effectiveness of our conflict modeling and pose refinement method through quantitative comparison demonstrated in Table \ref{tab:aba_mesh} and qualitative visualization shown in Figure \ref{fig:ablations_recon}. 
    As shown in Figures \ref{fig:ablations_recon} (a) and (b), the conflict modeling method effectively alleviates subtle inconsistencies in multi-view generation results, resulting in meshes with more realistic textures. 
    As visualized in Figures \ref{fig:ablation} (b) and (c), the regions with higher values in the conflict map correspond to the over-saturated and blurred areas in (b). 
    This indicates that our conflict map effectively captures inconsistencies in overlapping views, thereby enabling the generation of meshes with a high-fidelity texture.

\section {Conclusion}

     In this paper, we introduced \textit{NOVA3D}, a novel approach that unleashes 3D priors within a video diffusion model to generate high-quality textured meshes from any single image. By incorporating geometry information as an auxiliary supervisory signal and employing the Geometry-Temporal Alignment attention mechanism, our fine-tuned video diffusion model can generate dense-view aligned images and normal maps. 
     Furthermore, the de-conflict geometry fusion algorithm effectively resolves subtle multi-view conflicts in the generated images and addresses pose misalignments between the generated and pre-defined poses. Experimental results validate that our method delivers robust and generalizable performance, significantly outperforming existing baselines.

\bibliographystyle{IEEEbib}
\bibliography{NOVA3D}

\newpage % Ensures the appendix starts on a new page

\section*{\textbf{Appendix}}

\section{Training Details} \label{sec:Training}
    We start from the Stable Video Diffusion (SVD) model, which built on EDM-framewrok. We adopt preconditioning functions in the EDM-frameworks:
        \begin{equation}
            c_{skip}(\sigma) = \frac{\sigma ^2 _{data}}{\sigma^2 + \sigma_{data}^2}
        \end{equation}

        \begin{equation}
            c_{out}(\sigma) = \frac{\sigma  \sigma_{data}}{\sqrt{\sigma^2 + \sigma_{data}^2}}
        \end{equation}

    \begin{equation}
            c_{in}(\sigma) = \frac{1}{\sqrt{\sigma^2 + \sigma_{data}^2}}
        \end{equation}

        \begin{equation}
            c_{noise}(\sigma) = \frac{1}{4} ln(\sigma).
        \end{equation}
    We also adopt the noise distributuion and weighting function:
        \begin{equation}
            \mathrm{log}  \sigma \sim \mathcal{N} (P_{mean}, P^2_{std})
        \end{equation}
        \begin{equation}
           \lambda(\sigma) = (1 + \sigma^2) \sigma ^{-2}
        \end{equation}

    During training, we set \(\sigma_{\text{data}} = 1\), and progressively shift the noise distribution towards higher levels, which is found essential for high-quality video generation. Specifically, starting from the SVD pre-training configuration with \(P_{\text{mean}} = 1.0\) and \(P_{\text{std}} = 1.6\), we adjust the noise parameters to \(\{P_{\text{mean}}, P_{\text{std}}\} = \{1.8, 1.6\}\), \(\{2.2, 1.8\}\), and \(\{2.5, 2.0\}\) at 8,000, 16,000, and 24,000 global steps, respectively.
    
    Furthermore, unlike the multi-stage training strategy of Wonder3D~\cite{long2023wonder3d}, we employ a single-stage training approach after integrating SVD with the GTA attention mechanism. This ensures continuous information exchange between RGB and normal maps throughout training, thereby maximizing the retention of 3D priors from SVD. Our model is trained on our rendered multi-view dataset at a resolution of 256×256 using the AdamW optimizer with a learning rate of \(2 \times 10^{-5}\) in combination with exponential moving averaging at a decay rate of 0.9999 for approximately 30,000 steps.

\section{Implementation Details of GTA Module} \label{sec:GTA}

    To illustrate the proposed GTA attention mechanism, we detail the implementation of the basic RGB normal alignment attention in Algorithm 1. Additionally, we provide a detailed description of our GTA-infused Video Transformer Block in Algorithm 2.

    \begin{algorithm}   \label{alg:alignment_attention}
        \caption{Alignment Attention}
        \begin{algorithmic}[0]
            \STATE \textbf{Input:} $z$   {\quad// (nv   2 b )   d   c}
            \STATE $query, key, value \gets W^q(z), W^k(z), W^v(z) $
            \STATE \textcolor{blue}{// decomposition the rgb batch and normal batch}
            \STATE $key\_rgb, key\_norm \gets \mathrm{torch.chunk}(key)$
            \STATE $value\_rgb, value\_norm \gets \mathrm{torch.chunk}(value)$
            \STATE \textcolor{blue}{// concat rgb and normal latent on token length dim}
            \STATE $key \gets \mathrm{torch.cat}([key\_rgb, key\_norm], dim = 1)$
            \STATE $value \gets \mathrm{torch.cat}([value\_rgb, value\_norm], dim = 1)$
            \STATE $z \gets attention(key, vaule, query)$
            \RETURN $z$
        \end{algorithmic}
    \end{algorithm}

    \begin{algorithm}    \label{Video Transformer Block}
        \caption{GTA-Infused Video Transformer Block}
        \begin{algorithmic}[0]
        \STATE \textbf{Input:}   $z$, $embeddings$ for cross-attn
        \STATE \textcolor{blue}{// Spatial layer}
        % \STATE \textcolor{blue}{// hidden state shape: (nv   2b)   c   h   w}
        \STATE $z \gets \mathrm{ResBlock}(z)$
        \STATE $z \gets \mathrm{SelfAttention}(z)$
        \STATE \textcolor{blue}{// Frame Wise Alignment Attention}
        \STATE $z \gets \mathrm{AlignmentAttention}(z)$
        \STATE $z \gets \mathrm{CrossAttention}(z)$
        \STATE $z \gets \mathrm{Conv3D}(z)$
        \STATE \textcolor{blue}{// Temporal layer}
        \STATE $z \gets rearrange(z, (nv   2b)   c   h   w \to (2b   h   w)    nv
           c)$
        \STATE $z \gets \mathrm{ResBlock}(z)$
        \STATE $z \gets \mathrm{SelfAttention}(z)$
        \STATE \textcolor{blue}{// Temporal Wise Alignment Attention}
        \STATE $z \gets \mathrm{AlignmentAttention}(z)$
        \STATE $z \gets \mathrm{CrossAttention}(z)$
        \STATE $z \gets rearrange (z, ( 2b   h   w)    nv   c    \to    (nv   2b)   c   h   w) $

        \RETURN $z$
        \end{algorithmic}
    \end{algorithm}

    \FloatBarrier

    \section{Additional Results}
    \label{sec:additional results}
\begin{figure*}[hpt]
% \vspace{-2cm}
    \centering
    \includegraphics[width=\textwidth]{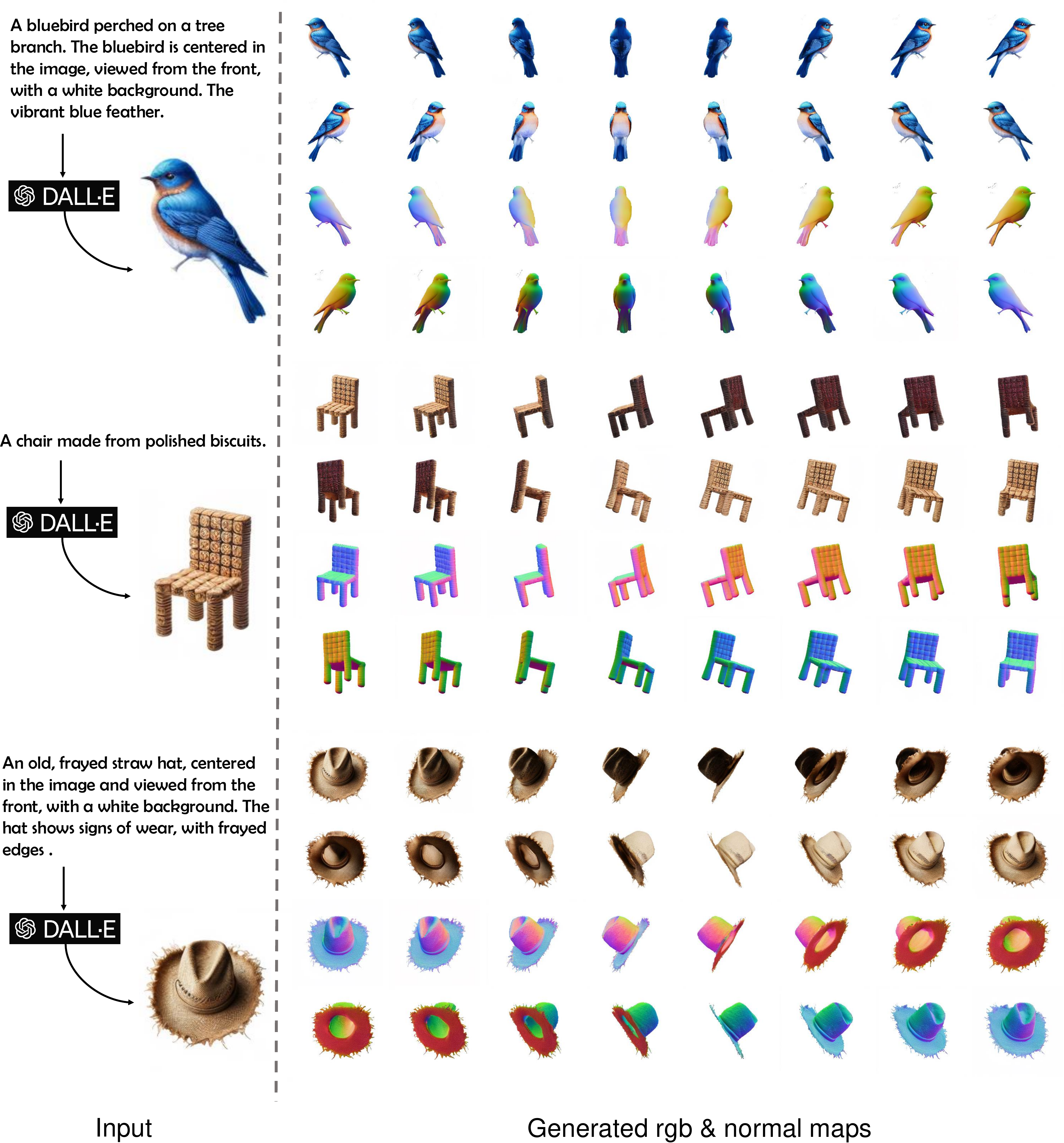}
    \caption{The qualitative results of NOVA3D on generated images and normal maps conditioned on in-the-wild images generated by of-the-shelf AIGC tool. }
    \label{fig:app_results}

\end{figure*}

    To evaluate the generalizability of our model, we utilize a 2D AI-generated content (AIGC) tool to create in-the-wild images, ensuring that these images are excluded from the training dataset.
    As shown in Figure \ref{fig:app_results}, \textit{NOVA3D} generates 16 views of RGB images and normal maps with remarkable cross-view consistency and a coherent overall shape, showcasing the strong generalization capabilities of our model.

    \clearpage

\end{document}